\PassOptionsToPackage{table}{xcolor}
\documentclass[authorversion,nonacm, %
     sigconf = true,
     screen = true,
     review = false,     %
     anonymous = false,  %
]{acmart}

\usepackage{soul}

\usepackage{booktabs} %
\usepackage{colortbl} %

\graphicspath{{pics/}}

\usepackage{etoolbox}
\usepackage{amsmath}
\usepackage{amsthm}
\usepackage{xspace}
\usepackage[table]{xcolor}
\usepackage{dsfont}
\usepackage{rotating}
\usepackage{bm}
\usepackage{subcaption}
\usepackage{hyperref}

\usepackage[algo2e,ruled,vlined,linesnumbered]{algorithm2e}
\usepackage{url}
\usepackage[noabbrev,nameinlink]{cleveref}

\crefname{prop}{property}{properties}
\creflabelformat{prop}{#2{\upshape(#1)}#3}

\crefname{pol}{policy}{policies}
\creflabelformat{pol}{#2{\upshape(#1)}#3}

\makeatletter
\def\NAT@spacechar{~}
\makeatother

\DeclareMathOperator*{\argmax}{arg\,max}

\newcommand{\R}{\mathbb{R}}
\newcommand{\N}{\mathbb{N}}

\newcommand{\rls}{\textsc{RLS}\xspace}

\newcommand{\onemax}{\textsc{OneMax}\xspace}

\newcommand{\leadingones}{\textsc{LeadingOnes}\xspace}
\newcommand*{\oporls}{\texorpdfstring{$(1 + 1)$}{(1 + 1)}~\rls}
\newcommand*{\oplrls}{\texorpdfstring{$(1 + \lambda)$}{(1 + lambda)}~\rls}

\DeclareMathOperator{\flip}{flip}

\DeclareMathOperator{\opt}{opt}

\newcommand{\K}{\mathcal{K}} %
\newcommand{\D}{\mathcal{D}} %

\newcommand{\optimal}{\texttt{optimal}\xspace}
\newcommand{\powersOfTwo}{\texttt{powers\_of\_2}\xspace}
\newcommand{\initialSegment}{\texttt{initial\_segment}\xspace}
\newcommand{\evenlySpread}{\texttt{evenly\_spread}\xspace}

\definecolor{teal}{RGB}{7, 171, 160}

\allowdisplaybreaks

\begin{document}

\title[Theory-inspired Parameter Control Benchmarks for Dynamic Algorithm Configuration]{Theory-inspired Parameter Control Benchmarks\\ for Dynamic Algorithm Configuration}

\author{André Biedenkapp$^*$}\thanks{$^*$equal contribution}
\affiliation{
\institution{University of Freiburg}
\city{Freiburg}
\country{Germany}
}

\author{Nguyen Dang$^*$}
\affiliation{
\institution{University of St Andrews}
\city{St Andrews}
\country{United Kingdom}
}

\author{Martin~S. Krejca$^*$}
\affiliation{
\institution{Sorbonne Universit{\'e}, CNRS, LIP6}
\city{Paris}
\country{France}
}

\author{Frank Hutter}
\affiliation{
\institution{University of Freiburg}
\city{Freiburg}
\country{Germany}
}
\affiliation{
\institution{Bosch Center for Artificial Intelligence}
\city{Renningen}
\country{Germany}
}

\author{Carola Doerr}
\affiliation{
\institution{Sorbonne Universit{\'e}, CNRS, LIP6}
\city{Paris}
\country{France}
}

\begin{abstract}
It has long been observed that the performance of evolutionary algorithms and other randomized search heuristics can benefit from a non-static choice of the parameters that steer their optimization behavior. Mechanisms that identify suitable configurations on the fly (``parameter control'') or via a dedicated training process (``dynamic algorithm configuration'') are therefore an important component of modern evolutionary computation frameworks. Several approaches to address the dynamic parameter setting problem exist, but we barely understand which ones to prefer for which applications. As in classical benchmarking, problem collections with a known ground truth can offer very meaningful insights in this context. Unfortunately, settings with well-understood control policies are very rare.

One of the few exceptions for which we know which parameter settings minimize the expected runtime is the LeadingOnes problem. We extend this benchmark by analyzing optimal control policies that can select the parameters only from a given portfolio of possible values. This also allows us to compute optimal parameter portfolios of a given size.
We demonstrate the usefulness of our benchmarks by analyzing the behavior of the DDQN reinforcement learning approach for dynamic algorithm configuration.
\end{abstract}

\maketitle

\section{Introduction}
\label{sec:intro}

It is well known that the performance of evolutionary algorithms and other black-box optimization heuristics can benefit quite significantly from a non-static choice of the (hyper-)parameters that determine their search behavior~\cite{SA83,HansenO01,battiti-book08,loshchilov-iclr17a,DoerrD18ga,BurkeGHKOOQ13,jaderberg-arxiv17a,parker-holder-neurips20}.
Not only does a dynamic choice of the parameters allow to tailor the search behavior to the specific problem instance at hand, but it can also be used to leverage complementarity between different search strategies during the different stages of the optimization process, e.g., by moving from a global to a local generation of solution candidates over the course of a search trajectory.

Mechanisms to identify suitable dynamic parameter values are intensively studied since decades, see~\cite{KarafotiasHE15,AletiM16,DoerrD18chapter} for surveys. Most works focus on generally applicable mechanisms to control the parameters \emph{on-the-fly}, e.g., using self-adaptation~\cite{Back98selfadaptation}, success-based parameter update strategies such as the one-fifth success rule~\cite{Rechenberg}, co-variance matrix adaptation~\cite{HansenO01}, or reinforcement learning~\cite{DaCostaGECCO08} (RL). However, for many practical applications of black-box optimization techniques we also have the possibility to \emph{learn} a parameter control policy via a dedicated training process, either because we anyway need to solve several instances of the same problem or because we can generate instances that are structurally similar to the ones that we expect to see in the future application. Our hope is then to derive structural insight into the algorithms' behavior that can be leveraged to choose their parameters in a more informed manner, just as we are used to do it for classic parameter tuning~\cite{Birattari09AlgorithmTuning,Hoos12AAC,SmitE09ComparingTuning}.

The study of parameter control schemes with dedicated offline training is recently enjoying growing attention in the broader AI community, where optimization heuristics are considered an interesting application of AutoML techniques~\cite{HutterKV19}. Examples include the training of a controller for the mutation strategy employed by differential evolution optimizing the CEC2015 problem collection~\cite{ManuelGECCO2019DDQN}
and learning to control the mutation step-size parameter of CMA-ES on the BBOB benchmarks~\cite{shala-ppsn20}.
The problem of training parameter control policies for strong performance on a distribution of instances was coined \emph{dynamic algorithm configuration (DAC)} in~\cite{BiedenkappBEHL20DACECAI}, where it is formulated as a contextual Markov Decision Process (see \Cref{sec:DAC} for details). To investigate the functioning and the performance of different DAC approaches, a dedicated library of benchmark problems, \emph{DACBench}, was suggested in~\cite{DACBench}.

With its rich history of parameter control studies, evolutionary computation has numerous exciting benchmark problems to offer for DAC, e.g., all the problems where dynamic parameter settings have been shown to outperform static ones. One such problem that is particularly well understood is the dynamic fitness-dependent selection of the mutation rates of greedy evolutionary algorithms maximizing the \leadingones problem (see \Cref{sec:generalSetup}).
In particular, we know exactly how the expected runtime of these algorithms depends on the mutation rates used during the run, and this is not only in asymptotic terms, but also for concrete problem dimensions $n$~\cite{BottcherDN10,Sudholt13,DoerrW18,Doerr19domi}. This feature has promoted \leadingones as an important benchmark for parameter control studies, both for empirical~\cite{DoerrDY16PPSN,DoerrW18} and for rigorously proven~\cite{LissovoiOW20,DoerrLOW18LO,DoerrDL21} results.

Our in-depth knowledge for \leadingones makes the problem an ideal candidate for
the in-depth empirical study of the pros and cons of DAC methods:
not only does the setting offer relatively fast evaluation times, but we also benefit from a ground truth against which we can compare the policies that are learned during the offline training phase.
Existing DAC benchmarks that give access to ground truth either abstract away the actual optimization process and replace it with a simple surrogate or they replace problem instances with unrealistic, artificial proxies.
Further, many traditional deep reinforcement learning benchmarks have deterministic environments, which makes them less representative for the configuration of metaheuristics.
\leadingones can therefore fill an important gap.

\textbf{Our Contributions.}
We demonstrate in this work how the mutation control problem for \leadingones can be used to investigate existing DAC approaches and their capabilities.
We exemplarily evaluate a commonly used reinforcement learning approach using deep neural networks (dubbed DDQN) and investigate how it scales with different problem dimensions.

Each problem dimension of \leadingones provides us with a different problem instance on which we can compare the results of the DAC process to the \emph{ground truth,} i.e., the optimal strategy.\!\footnote{\label{foot:runtime}All optimality claims made here and in the remainder of the paper are always with respect to expected runtime. This is also our primary performance measure, i.e., when we speak of the \emph{performance} of an algorithms, we refer to the expected number of fitness evaluations made before an optimal solution is evaluated for the first time.}
To enrich the problem collection further, we also compute optimal control policies for settings in which the algorithms are only allowed to select their parameter values from a given portfolio $\K$ of possible values (\Cref{tab:optimal_policies}). These results generalize previous works of \citet{LissovoiOW20}, who analyzed optimal policies for the portfolios that are composed of the integers $i \in [1, k] \cap \N$ for $k \in \Theta(1)$.

We observe that for smaller settings, both in terms of problem size $n$ and portfolio size $k$, the employed DAC method is capable of learning optimal policies quickly (\Cref{sec:50}).
Increasing either $n$ or $k$, however, can drastically increase the difficulty for the learning method, resulting in potentially sub-optimal policies or even no successful learning within the given budget and hyperparameters setting (\Cref{fig:performance_evenly_spread}).

Of independent interest for the runtime analysis community are the optimal parameter portfolios (\Cref{tab:optimal_portfolios}) that we compute for a number of different combinations of problem dimension $n$, and portfolio size $k$. While these optimal portfolios have a large intersection with the \initialSegment portfolio investigated by \citet{LissovoiOW20}, the optimal performance achieved with this portfolio is worse than the performance achieved with the portfolio of exponentially growing values $\{2^i \mid i \in [0, k-1] \cap \N\}$.

\textbf{Outline.}
In \Cref{sec:generalSetup}, we introduce our benchmark, consisting of the \leadingones problem as well as the \oporls algorithm.
In \Cref{sec:theory}, we explain how to derive optimal policies for a given portfolio.
Further, we analyze these policies with respect to increasing portfolio and dimension size.
In \Cref{sec:RL}, we analyze empirically how well optimal policies can be learned when using the DDQN reinforcement learning approach.
Like in \Cref{sec:theory}, we consider different portfolios as well as increasing portfolio and dimension sizes.
Last, we conclude our work in \Cref{sec:conclusion}.

\textbf{Availability of Code and Data.} Our implementations and our results are available on GitHub at~\cite{data}.

\section{Parametrized RLS for LeadingOnes}
\label{sec:generalSetup}

We consider the optimization of the \leadingones problem via variants of randomized local search, which we present in the following.
We note that we use, for all $a, b \in \N$, the notation $[a .. b] \coloneqq [a, b] \cap \N$.

\begin{algorithm2e}[t]%
    \caption{
        The \oporls with state space~$\mathcal{S}$, portfolio $\K \subseteq [0 .. n]$, and parameter selection policy $\pi\colon\mathcal{S} \to \K$, maximizing a function $f\colon \{0,1\}^n \to \R$.
        See also \Cref{sec:generalSetup}.
    }
    \label{alg:rls}
   	$x \gets$ a sample from $\{0,1\}^{n}$ chosen uniformly at random\;
    \For{$t \in \N$}{
        $s \gets$ current state of the algorithm\;\label{line:getState}
        $r \gets \pi(s)$\;
        $y \gets \flip_{r}(x)$\;
   		\lIf{$f(y)\geq f(x)$}{$x \leftarrow y$}
   	}
\end{algorithm2e}

\textbf{Parameterized Randomized Local Search.}
We analyze a parameterized version of the classic randomized local search (\rls) algorithm. While RLS searches only in the direct neighborhood of a current-best solution, its parametrerized cousin, the \oporls (\Cref{alg:rls}), can sample solution candidates at larger distances.

The \oporls maintains a single bit string (the \emph{current solution}), denoted by~$x$ in \Cref{alg:rls}, initially drawn uniformly at random from $\{0, 1\}^n$.
Iteratively, the \oporls generates a new sample~$y$ (the \emph{offspring}) from the current solution $x$, and it replaces $x$ with~$y$ if the the objective value $f(y)$ (its \emph{fitness}) is at least as large as $f(x)$.
The offspring~$y$ is generated by the operator $\flip_{r}$ (the \emph{mutation}), which, given a parameter $r \in [0 .. n]$, inverts exactly~$r$ pairwise different bits in~$y$, chosen uniformly at random from all possible $r$-subsets of the index set $[1 .. n]$.
We call the parameter~$r$ of the mutation the \emph{search radius}.
In each iteration, the \oporls chooses the search radius to apply based on a function~$\pi$ that we call a \emph{(parameter selection) policy}, given some state of the algorithm.
The policy~$\pi$ only returns search radii from a certain set~$\K \subseteq [0..n]$, which we call the \emph{portfolio} of the algorithm.
Note that the portfolio~$\K$ and the policy~$\pi$ are part of the input of the \oporls.

Although rich states can prove useful~\cite{BuzdalovB15CEC}, we only have theoretical guarantees for \emph{fitness-dependent} policies, which use exclusively the fitness of the current solution.
\citet{DoerrL18LO} discuss why it is hard to derive more general bounds.
Thus, we assume in this article that the policies are \emph{fitness-dependent}.

As mentioned in~\Cref{foot:runtime}, our key performance criterion is the number of iterations until the \oporls finds a global optimum of its fitness function for the first time, that is, the smallest $t \in \N$ such that~$x$ is optimal at the beginning of that iteration.
We refer to this number as the algorithm's \emph{runtime}, noting that it is a random variable.

\textbf{\leadingones.}
The \leadingones problem is defined over bit strings of length $n \in \N$. It asks to maximize the number of leading~$1$s of a bit string.
Formally, $\leadingones\colon \{0,1\}^n \to [0 .. n], x \mapsto \max\{i \in [0..n] \mid \forall j \le i\colon x_j = 1\}$.
Note that the unique global maximum is the all-$1$s string.

\leadingones is a special case of the general problem of maximizing the longest prefix of agreement with a hidden target bit string $z \in \{0,1\}^n$, evaluated with respect to a hidden permutation~$\sigma$ that shuffles the bit positions, formally defined as $\leadingones_{z, \sigma}\colon\newline \{0,1\}^n \to [0 .. n], x \mapsto \max\{i \in [0..n] \mid \forall j \le i\colon x_{\sigma(j)} = z_{\sigma(j)}\}$.
Since the \oporls is unbiased in the sense of \citet{LehreW12}, its performance is identical on each of these problem instances and we therefore restrict our attention to the classic $\leadingones$ instance mentioned above.

Although \leadingones can be solved using $\Theta(n \log\log n)$ queries in expectation~\cite{AfshaniADDLM19}, this runtime cannot be achieved with unary unbiased algorithms such as the \oporls.
Their runtime grows at least quadratically in the dimension~\cite{LehreW12}.
The same bound of $\Omega(n^2)$ also applies to all (1+1) elitist algorithms~\cite{DoerrL18LO}, of which the \oporls is a representative as well.
The expected runtime of the classic \rls with constant search radius 1 is $n^2/2$.

\section{Optimal Policies and Portfolios for \leadingones}
\label{sec:theory}
The exact runtime distribution for \leadingones is well understood for the \oporls~\cite[Section~$2.3$]{Doerr19domi}.
Its expected runtime is, besides its initialization, entirely determined by how quickly it improves the fitness of its current solution.
More formally, the most important values are the~$n$ different probabilities $(p_i)_{i \in [0..n - 1]}$, where, for each $i \in [0..n - 1]$, the value~$p_i$ denotes the probability that the \oporls finds a strict improvement if the current solution has fitness~$i$.
Choosing for each $i \in [0..n - 1]$ the search radius so that~$p_i$ is maximized results in an \oporls instance with optimal runtime on \leadingones.

In more detail, for each $i \in [0..n - 1]$ and each $r \in [0 .. n]$, let $q(r, i)$ denote the probability that the \oporls finds a strict improvement if the current solution has fitness~$i$ and flips exactly~$r$ bits during mutation.
For \leadingones, it holds for all $i \in [0..n - 1]$ and all $r \in [0 .. n]$ that~\cite[Section~$2.3$]{Doerr19domi}
\begin{align}
    \label{eq:flipping_probability}
    q(r, i) = \frac{r}{n} \cdot \prod\nolimits_{j \in [1 .. r - 1]} \frac{n - i - j}{n - j}.
\end{align}

An important property of~$q$ that allows to determine optimal policies for various portfolios of the \oporls is that, for all $i \in [0..n - 1]$ and $r \in [0 .. n - 1]$, it holds that~\cite[Section~$2.3$]{Doerr19domi}
\begin{align}
    \label[prop]{prop:monotonicity_property}
    q(r, i) \leq q(r + 1, i) \textrm{ if and only if } i \leq (n - r)/(r + 1).
\end{align}

In the following, in \Cref{sec:unbounded}, we discuss what an optimal policy looks like for the well understood case when permitting \emph{all} possible search radii from~$0$ to~$n$.
We refer to this setting as the \emph{full portfolio}.
Afterward, we explain in \Cref{sec:bounded} how to calculate optimal policies when the portfolio does not contain all search radii, that is, when it is \emph{restricted}.
Last, in \Cref{sec:compare_optimal_policies}, we compare optimal policies of different portfolios, including the optimal one, which, given a portfolio size and a problem dimension, minimizes the expected runtime.

\textbf{Generalizations.}
We note that our analyses can easily be extended to the \oplrls, the variant of the \oporls that generates $\lambda \in \N_{\geq 1}$ offspring in each iteration.
For this algorithm, \cref{eq:flipping_probability} looks slightly different, as it incorporates~$\lambda$, but all other arguments work out in the same way.

\subsection{Full Portfolio}
\label{sec:unbounded}
In the setting of $\K = [0..n]$, an optimal policy $\pi_{\opt}$ satisfies~\cite{DoerrW18,Doerr19domi}
\begin{align}
    \label[pol]{pol:optimal_policy_full_portfolio}
    \pi_{\opt}\colon i \mapsto \lfloor n/(i+1) \rfloor.
\end{align}
This is a direct consequence of \cref{prop:monotonicity_property}, as it can be proven that this is the policy that chooses for each fitness value $i \in [0..n-1]$ the $r \in [0 .. n]$ that maximizes $q(r, i)$.

Note that \cref{pol:optimal_policy_full_portfolio} is monotonically decreasing.
That is, the higher the fitness of the current individual, the fewer bits are flipped.
This entails that not all search radii are used.
For example, for a fitness of~$0$, it is optimal to flip all~$n$ bits.
For a fitness of~$1$, it is optimal to flip exactly $\lfloor n/2 \rfloor$ bits.
Thus, $\pi_{\opt}$ skips over all search radii in the range $[\lfloor n/2 \rfloor + 1 .. n - 1]$.
We further note that using $\pi_{\opt}$ results in an expected runtime of about $0.39 n^2$ on \leadingones and that using only the search radius~$1$ results in an expected runtime of $0.5 n^2$~\cite[Section~$2.3$]{Doerr19domi}.
Thus, the expected runtime of \emph{any} portfolio with search radius~$1$, using an optimal policy, falls into this range.

\subsection{Restricted Portfolio Sizes}
\label{sec:bounded}
For $\K \subsetneq [0..n]$, the optimal policy $\pi_{\opt}^{(\K)}$ strongly depends on the search radii in~$\K$.
Thus, in general, the policy cannot follow an easy formula as given by~$\pi_{\opt}$ in \cref{pol:optimal_policy_full_portfolio} but needs to be adjusted to the specific values available in $\K$.
Further, if $1 \notin \K$, then the expected runtime of an algorithm using~$\K$ can be infinite (in particular when the probability of creating a solution with fitness $n-1$ is non-zero, as such a solution can only be improved with search radius~$1$).
Thus, we assume in the following always that $1 \in \K$.

\subsubsection{Determining an optimal policy}
\label{sec:determine_optimal_policy}
Let $i \in [0 .. n - 1]$ denote the fitness of the current individual, and assume that $\pi_{\opt}(i) \notin \K$.
Due to \cref{prop:monotonicity_property}, the best possible search radius in~$\K$ is one of the at most two values closest to~$\pi_{\opt}(i)$.
In other words, $\pi_{\opt}^{(\K)}(i)$ is either $r^{\sup}_i \coloneqq \max\{r \in \K \mid r < \pi_{\opt}(i)\}$ or $r^{\inf}_i \coloneqq \max\{r \in \K \mid r > \pi_{\opt}(i)\}$.
Thus, it holds that
\begin{align}
    \label{pol:optimal_policy_restricted_portfolio}
    \pi_{\opt}^{(\K)}(i) = \argmax\nolimits_{r \in \{r^{\sup}_i, r^{\inf}_i\}} q(r, i).
\end{align}
Note that this implies that~$\pi_{\opt}^{(\K)}$ is monotonically decreasing, as, for all $i, j \in [0 .. n - 1]$, $i < j$, it holds that $r^{\sup}_i \geq r^{\sup}_j$ and $r^{\inf}_i \geq r^{\inf}_j$.

Let~$\D$ denote the vector of the elements of~$\K$ in decreasing order.
The monotonicity of \cref{pol:optimal_policy_restricted_portfolio} allows to simplify the calculations for~$\pi_{\opt}^{(\K)}$ by only determining the fitness values for which the the probability of improvement~$q$ for two consecutive elements in~$\D$ changes.
That is, we only need to determine for all $i \in [1 .. |\K| - 1]$ the largest $j \in [0 .. n]$ such that $q(\D_i, j) \geq q(\D_{i + 1}, j)$.
We call each of these $|\K| - 1$ points~$j$ a \emph{breaking point}.
We note that breaking points do not need to be unique.
\Cref{alg:determine_optimal_breaking_points} provides a pseudo code for how to determine the breaking points for a given portfolio~$\K$.
Note that \cref{line:linear_search_loop,line:linear_search_condition,line:linear_search_update} can be improved by applying a binary search that returns the smallest index at which the condition from \cref{line:linear_search_condition} holds.
This is avoided here in favor of simplicity.

\begin{algorithm2e}[t]%
    \caption{
        The algorithm to compute, for a given portfolio~$\K$ with $1 \in \K$ the breaking points $(b_i)_{i \in [1 .. |\K| - 1]}$ of the optimal policy~$\pi_{\opt}^{(\K)}$, as discussed in \Cref{sec:bounded}.
        The function~$q$ is defined in \cref{eq:flipping_probability}.
    }
    \label{alg:determine_optimal_breaking_points}
    $\D \gets \K$ in descending order\;
    $c \gets 0$\;
    \For{$i \in [1 .. |\K| - 1]$}%
    {%
        \For{$j \in [1 .. n]$\label{line:linear_search_loop}}%
        {%
            \lIf{$q(\D_i, j) < q(\D_{i + 1}, j)$\label{line:linear_search_condition}}%
            {%
                break the loop over~$j$%
            }
            $c \gets j$\;\label{line:linear_search_update}
        }
        $b_i \gets c$\;
    }
\end{algorithm2e}

Given the breaking points $(b_i)_{i \in [1 .. |\K| - 1]}$ of a portfolio~$\K$ and defining $b_0 = -1$ and $b_{|\K|} = n - 1$, the optimal policy~$\pi_{\opt}^{(\K)}$ is easily calculated by noting that, for all $i \in [0 .. |\K|]$ and all $j \in [b_i + 1 .. b_{i+1}]$, the $i$-th largest value in $\K$ is the optimal search radius when the current individual has fitness~$j$.

\subsection{Comparing Optimal Policies}
\label{sec:compare_optimal_policies}
We compare different portfolios of the same size~$k$, and we compare their resulting optimal policies calculated as stated at the end of \Cref{sec:determine_optimal_policy}.
To this end, we consider the following four portfolios.
For $n \in \N_{\geq 2}$ and $k \in [2 .. n]$, we define
\begin{itemize}
    \item {\powersOfTwo} to be $\{2^i \mid 2^i \leq n \land i \in [0 .. k - 1]\}$,
    \item {\initialSegment} to be $[1 .. k]$,
    \item {\evenlySpread} to be $\{i \cdot \lfloor n/k \rfloor + 1 \mid i \in [0 .. k - 1]\}$, and
    \item {\optimal}, which we determine by a brute-force approach over all $k$-subsets of~$n$ that contain the search radius~$1$.
        The portfolio with the lowest expected runtime among all of these subsets is considered \optimal.
\end{itemize}
Note that \powersOfTwo is only defined for values~$k$ of at most $\lfloor \log_2 n \rfloor$.
For any larger value of~$k$, it is not defined.
Last, note that although there is only one \optimal \emph{portfolio}, all \emph{policies} discussed in this section are optimal with respect to their specified portfolio.

\textbf{The portfolio \optimal.}
\Cref{tab:optimal_portfolios} shows optimal portfolios for $n \in \{50, 100\}$ and for $k \in [2 .. 8]$.
For these cases, the portfolio consists of the interval $[1 .. \lceil k/2 \rceil]$ and of some larger values that seem to grow exponentially.
That is, \optimal is a mixture of \initialSegment and a variant of \powersOfTwo.
Interestingly, for $k = 8$, the portfolio contains the search radius $50 = n$, which is only relevant if the current individual of an algorithm has a fitness of~$0$.
Due to the uniform initialization, the probability that we see this value is $50\,\%$, and we transition to a different state with probability 1 by flipping all bits, so that the difference between the optimal expected runtime that can be achieved with a portfolio of size $k=8$ over that for $k=7$ is at most $0.5$.
Further, \optimal is identical for $n \in \{50, 100\}$ for $k \in \{2, 3, 4\}$.
For larger~$k$, some larger search radii change slightly.
This suggests that the generals range of optimal search radii to use is only slightly affected by the problem size.

\begin{table}[t]
    \caption{
        The optimal portfolios for various sizes~$k$, for the problem sizes $n \in \{50, 100\}$, and their expected runtimes (divided by~$n^2$).
        For $k = 8$ and $n = 100$, the computation took too long.
        See also \Cref{sec:compare_optimal_policies}.
    }
    \label{tab:optimal_portfolios}
    \begin{tabular}{rllll}
            & \multicolumn{4}{c}{Optimal portfolio/Expected runtime by $n^2$}                  \\
        $k$ & \multicolumn{2}{l}{$n = 50$} & \multicolumn{2}{l}{$n = 100$}                     \\
        \toprule
        $2$ & $1, 4$                     & $0.409832$  & $1, 4$                  & $0.409897$  \\
        $3$ & $1, 2, 6$                  & $0.39568$   & $1, 2, 6$               & $0.395987$  \\
        $4$ & $1, 2, 4, 11$              & $0.3911372$ & $1, 2, 4, 11$           & $0.391403$  \\
        $5$ & $1, 2, 3, 6, 17$           & $0.3895904$ & $1, 2, 3, 6, 16$        & $0.389892$  \\
        $6$ & $1, 2, 3, 5, 9, 21$        & $0.3888308$ & $1, 2, 3, 5, 9, 23$     & $0.389109$  \\
        $7$ & $1, 2, 3, 4, 6, 12, 29$    & $0.388452$  & $1, 2, 3, 4, 6, 11, 27$ & $0.3887584$ \\
        $8$ & $1, 2, 3, 4, 6, 9, 19, 50$ & $0.3882052$ & --                      & --
    \end{tabular}
\end{table}

\textbf{Optimal policies.}
\Cref{tab:optimal_policies} shows optimal policies (depicted as their relative breaking points) for different portfolio sizes~$k$ and problem dimensions~$n$.
For \powersOfTwo and \initialSegment, when increasing~$k$, the portfolio is extended by adding larger search radii.
This is reflected in their respective (optimal) portfolio, as the breaking points are also extended.
In contrast, for \evenlySpread, a portfolio of one size is \emph{not} an extension of a portfolio of a smaller size.
This is reflected in the breaking points, which are not extended for increasing~$k$.
For all cases of~$n$ and~$k$ depicted, \powersOfTwo and \initialSegment share at least half of their breaking points with \optimal.
This follows also from the results of \Cref{tab:optimal_portfolios}, which shows that the high overlap of \optimal with \initialSegment continues, whereas the one with \powersOfTwo is not that prominent for larger~$k$.
Since all portfolios except for \evenlySpread contain at least the search radii~$1$ and~$2$, the optimal policies also utilize the full range of these radii, following \cref{pol:optimal_policy_full_portfolio}.
For \evenlySpread, mostly the search radius~$1$ is important.

\begin{figure}
    \includegraphics[width=0.9\columnwidth]{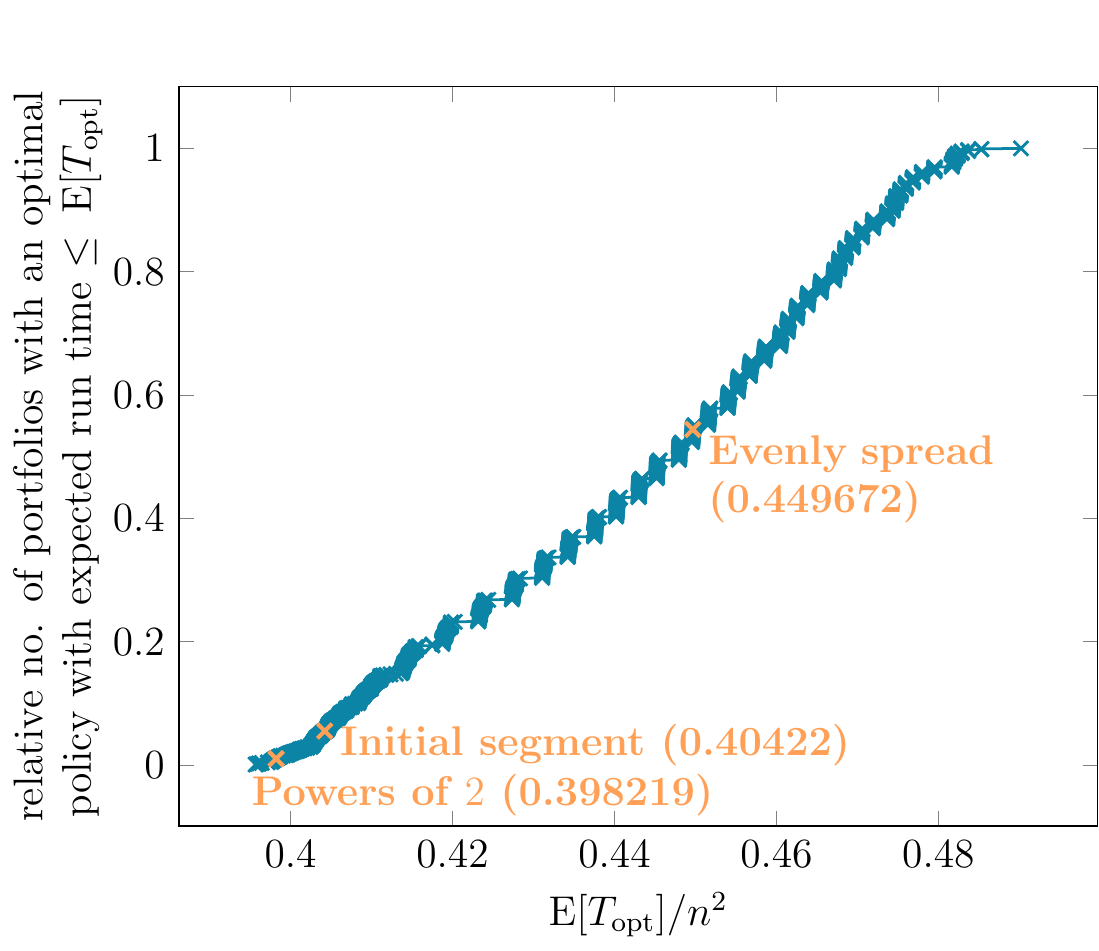}
    \caption{
        The cumulative fraction of how many portfolios among \emph{all} portfolios have at most the expected (relative) runtimes stated by the $x$-axis, for $n = 50$.
        All portfolios have a cardinality of exactly~$3$ and contain the search radius~$1$.
        Their expected runtime is determined by applying an optimal policy.
        See also \Cref{sec:compare_optimal_policies}.
    }
    \label{fig:cumulative_policy_plot}
    \Description[Short description.]{Long description.}
\end{figure}

\Cref{fig:cumulative_policy_plot} investigates the case of $k = 3$ for $n = 50$ more closely.
We computed for all $\binom{50}{2}$ portfolios of size~$3$ that contain the search radius~$1$ the expected runtime of an optimal policy.
The figure depicts cumulative data of these computations.
Interestingly, the curve follows an almost linear trend, except for the last $5$\,\%, where the increase in the expected runtime is diminishing.
This suggests that choosing portfolios uniformly at random has a fair chance of resulting in a good expected runtime of its optimal policy.

\begin{figure}
    \includegraphics[width=0.9\columnwidth]{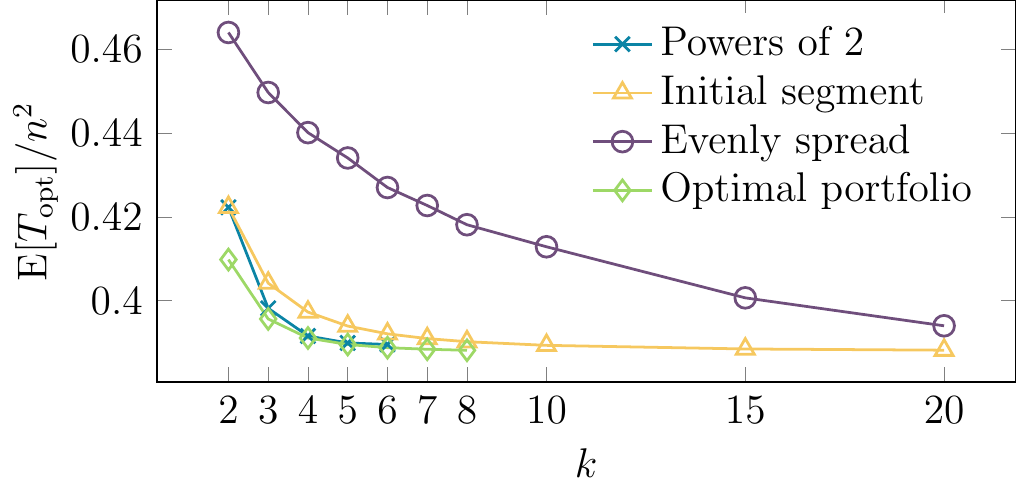}
    \caption{
        The expected runtimes for the optimal policies of the stated portfolios for $n = 50$.
        The runtime is divided by~$n^2$.
        See also \Cref{sec:compare_optimal_policies}.
        Note that \powersOfTwo is not defined for $k > 6$.
        Further, we only computed \optimal up to $k = 8$.
    }
    \label{fig:optimal_policies_comparison}
    \Description[Short description.]{Long description.}
\end{figure}

In \Cref{fig:optimal_policies_comparison}, we take a closer look at the impact of the portfolio size~$k$ on the expected runtime.
The figure compares the expected runtimes of all four different portfolios defined above when using an optimal policy.
Interestingly, although \initialSegment shares a large part of its search radii with \optimal (\Cref{tab:optimal_portfolios}), the expected runtime of \powersOfTwo is better than that of \initialSegment.
This suggests that having \emph{some} larger search radii is more beneficial than covering exclusively small search radii.
However, the comparably bad expected runtime of \evenlySpread shows that having more than a single small search radius (for example, $1$ \emph{and}~$2$) drastically improves the expected runtime.

\begin{table}[t]
    \caption{
        The breaking points (\Cref{alg:determine_optimal_breaking_points}) of different portfolios (\Cref{sec:compare_optimal_policies}) of size $k \in \{3, 4\}$ for $n \in \{50, 100\}$.
        Each breaking point is divided by~$n$.
        Recall that the breaking points refer to the portfolio sorted in descending order.
    }
    \label{tab:optimal_policies}
    \begin{tabular}{rlll}
        $k$ & Portfolio       & $n = 50$                 & $n = 100$                \\
        \toprule
        $3$ & \optimal        & $0.22, 0.48$       & $0.23, 0.49$       \\
            & \powersOfTwo    & $0.26, 0.48$       & $0.28, 0.49$       \\
            & \initialSegment & $0.3, 0.48$        & $0.32, 0.49$       \\
            & \evenlySpread   & $0, 0.12$          & $0, 0.08$          \\
        \midrule
        $4$ & \optimal        & $0.1, 0.26, 0.48$  & $0.12, 0.28, 0.49$ \\
            & \powersOfTwo    & $0.14, 0.26, 0.48$ & $0.15, 0.28, 0.49$ \\
            & \initialSegment & $0.22, 0.3, 0.48$  & $0.24, 0.32, 0.49$ \\
            & \evenlySpread   & $0, 0.02, 0.16$    & $0, 0.01, 0.1$     \\
    \end{tabular}
\end{table}

\section{Algorithm Configuration with Reinforcement Learning}
\label{sec:RL}

Parameter control with a dedicated offline training phase has long been studied~\cite[see e.g.,][]{ManuelGECCO2019DDQN,KarafotiasSE12,KeeAdaptiveGA,BurkeGHKOOQ13,VermettenCMAdynAS}.
Recently it gained attention in the broader AI community where \emph{dynamic algorithm configuration} (DAC) \cite{BiedenkappBEHL20DACECAI} was proposed as a generalization over algorithm configuration \cite{hutter-jair09a} and algorithm selection \cite{rice76a}.
In DAC, reinforcement learning (RL) is predominantly used to learn dynamic configuration policies.
In the DAC setting, our proposed benchmark is of particular interest as it readily allows us to investigate important questions such as:
i) Can DAC learn optimal policies?
ii) How does the choice of elements of the portfolio~$\mathcal{K}$ influence the learning procedure?
iii) How does the size of~$\mathcal{K}$ influence the learning procedure?
iv) How does the problem size influence the learning procedure?

We recap the most important definitions for DAC in \Cref{sec:DAC}. The experimental setup of our work is summarized in \Cref{sec:setup}. Results for small portfolios $|\K| \in \{3,4,5\}$ and for fixed dimension $n=50$ are presented in \Cref{sec:50} and results for broader ranges of portfolio sizes and dimensions are discussed in \Cref{sec:dim}.

\subsection{The DAC Framework}
\label{sec:DAC}

The process of dynamically adapting hyperparameters is modeled as a contextual Markov Decision Process (cMDP) \cite{hallak-corr15}.
An MDP $\mathcal{M}$ is a tuple $\langle \mathcal{S}, \mathcal{A}, \mathcal{T}, \mathcal{R}\rangle$ with state space $\mathcal{S}$, action space $\mathcal{A}$,  transition function $\mathcal{T}\colon\mathcal{S}\times\mathcal{A}\times\mathcal{S}\to[0,1]$ and reward function $\mathcal{R}\colon\mathcal{S}\times\mathcal{A}\to\mathbb{R}$.
The transition function describes the dynamics of the process and gives the probability of reaching a state $s'$ when playing action $a$ in state $s$.
Similarly, the reward function describes the reward obtained by playing action $a$ in $s$.
Depending on the system an MDP describes, the reward function can be stochastic.
A cMDP extends this formalism through the use of so-called \emph{context information} $i\sim\mathcal{I}$.
The context influences the behavior of the reward and transition functions but leaves the state and action spaces unchanged.
Thus a cMDP $\mathcal{M}=\{\mathcal{M}_i\}_{i\sim\mathcal{I}}$ is a collection of MDPs with shared state and action spaces, but with individual transition and reward functions ($\mathcal{T}_i$ and $\mathcal{R}_i$).
In DAC, the state space describes the internal behavior of an algorithm $A$ (e.g., internal statistics of $A$) when running it on an instance $i$ (i.e., the context) and the action space is given by the possible values of parameters of $A$.
In practice, the transition and reward functions are unknown and not trivial to approximate or learn.
Still, there exist solution approaches for MDPs that do not need direct access to these.

Reinforcement learning (RL) \cite{suttonbook} has been demonstrated to be able to learn dynamic configuration policies directly from data \citep[see e.g.,][]{lagoudakis-icml00,lagoudakis-endm01a,pettinger-gecco02,DaCostaGECCO08,sakurai-sitis10a,battiti-as12a,daniel-aaai16,ManuelGECCO2019DDQN,BiedenkappBEHL20DACECAI}. In an offline learning phase, an RL agent interacts with its environment (i.e., the algorithm that is being configured) to learn which actions lead to the highest reward over multiple episodes (trajectory until a goal state or a maximal step-limit is reached). In a trial-and-error fashion, an RL agent iteratively observes the current state $s_t$ of the environment at time $t$. Based on this observation it selects an action $a_t$ which advances the environment to the next state $s_{t+1}$ and produces a reward signal $r_{t+1}$.
This information is sufficient to learn the value of each state and how to select the next action to maximize the expected reward.

In the commonly used $\mathcal{Q}$-learning approach \cite{watkins89} the goal is to learn the $\mathcal{Q}$-function $\mathcal{Q}\colon\mathcal{S}\times\mathcal{A}\to\mathcal{R}$ that maps a state--action pair to the cumulative future reward that is received after playing an action $a$ in state $s$.
The $\mathcal{Q}$-function can be learned in a typical error correction fashion.
Given a state $s_t$ and action $a_t$, the $\mathcal{Q}$-value $\mathcal{Q}(s_t, a_t)$ can be updated using temporal differences (TD) as
\begin{equation*}
    \mathcal{Q}(s_t, a_t) \gets \mathcal{Q}(s_t, a_t) +  \alpha\Big(\underbrace{\big(\overbrace{r_{t} + \gamma \max\mathcal{Q}(s_{t+1}, \cdot)}^\text{TD-target}\big) - \mathcal{Q}(s_t, a_t)}_\text{TD-delta}\Big)
\end{equation*}
where $\alpha$ is the \emph{learning rate} and $\gamma$ is the \emph{discounting factor}.
The TD-target is the reward $r_t$ incurred by playing $a_t$ in $s_t$ together with the discounted maximal future reward.
The discounting factor determines how important future rewards are when updating the $\mathcal{Q}$-function.
The TD-delta then describes how correct or wrong the prediction was and is used to update the $\mathcal{Q}$-function accordingly.
The learning rate determines the strength with which the TD-delta updates the original prediction.
A reward-maximizing policy can then be defined by only using the learned $\mathcal{Q}$-function as
$
    \pi(s) = \argmax_{a\in\mathcal{A}} \mathcal{Q}(s,\cdot)
$.
For better exploration while learning, typically $\epsilon$-greedy exploration is used, where $\epsilon$ gives the probability that an action $a_t$ is replaced with a randomly sampled one.

\citet{mnih-nature13} proposed to model the $\mathcal{Q}$-function as a neural network (referred to as deep $\mathcal{Q}$-network) and showed that this allowed to learn $\mathcal{Q}$-functions even for high-dimensional states such as frames of video games.
\Citet{hasselt-aaai16a} showed that using a single network when selecting the maximizing action in the TD-target and in the prediction of the value often leads to instabilities due to overestimation during training.
To mitigate this, they proposed to use a second copy of the weights of the neural network. One set is used to select the maximizing action and the other is used to predict the value.
The second set of weights is kept frozen for short periods at a time and then copied over from the first set for increased stability of predictions.
This extension is dubbed double deep $\mathcal{Q}$-network (DDQN) and generally results in overall faster learning due to less overestimation. DDQN has been used as solution approach to DAC problems in DE~\cite{ManuelGECCO2019DDQN} and AI planning~\cite{speck-icaps21}.

\subsection{Experimental Setup}
\label{sec:setup}

Following \citet{BiedenkappBEHL20DACECAI}, in our experiments we use a small DDQN with two hidden layers and $50$ units each to learn the $\mathcal{Q}$-function. The action space $\mathcal{A}$ is the portfolio $\mathcal{K}$. We define $s_t=f(x_t)$ and $r_t=f(x_t)-f(x_{t-1}) - 1$, where $x_t$ is the solution found by the \oporls at time step $t$.
During the training of DDQN, we impose a cutoff time of $0.8n^2$ steps per episode to avoid wasting too much time sampling with bad policies. Recall that the expected run time of the simple setting with a constant policy  $\pi\colon s \mapsto 1$ is $0.5n^2$~\cite{Doerr19domi}. The episode-cutoff time for our RL training is chosen such that policies slightly worse than this trivial constant policy can still be explored during the learning phase.
All DDQN agents are trained with a batch size of $2048$, an $\epsilon$-greedy value of $0.2$, and a discount factor $\gamma$ of $0.9998$.
The batch size determines how many samples are used to compute the gradients when updating the neural network. A larger batch size results in a more accurate estimation of the gradient but takes longer to compute.

It is known that hyperparameters play a crucial role in deep RL algorithms~\cite{henderson-aaai18a}.
Tuning them is expensive and not trivial and many purpose-built methods exist depending on the target application and algorithm~\cite{AutoRLsurvey}.
It is, however, not well understood how the hyperparameters influence the learning behavior of agents, especially outside of the domain of video game playing.
We built our choice of hyperparameters on prior literature using RL for dynamic tuning and adjusted batch size and $\gamma$ based on results of a small prestudy.

\subsection{Results for \texorpdfstring{$n=50$}{n=50}}
\label{sec:50}
In the first set of experiments, we consider a fixed problem size of $n=50$ as well as the three portfolio settings \initialSegment, \powersOfTwo, and \evenlySpread from \Cref{sec:compare_optimal_policies}. For each setting, three portfolio sizes $k \in \{3,4,5\}$ are considered. The aim is to study the impact of portfolio settings and portfolio sizes on DDQN's learning behaviors. For each pair of portfolio settings and sizes, a DDQN agent is trained with a budget of 1 million time steps and a walltime limit of 24 hours on an $8$-core Intel Xeon E5-4650L computer ($2.6$\,GHz). The best policy is chosen at the end of the training phase
and
is then evaluated and compared against the optimal policy of the same portfolio $\mathcal{K}$ via $2000$ runs (per policy).

\begin{figure}[t]
    \centering
    \includegraphics[width=\columnwidth]{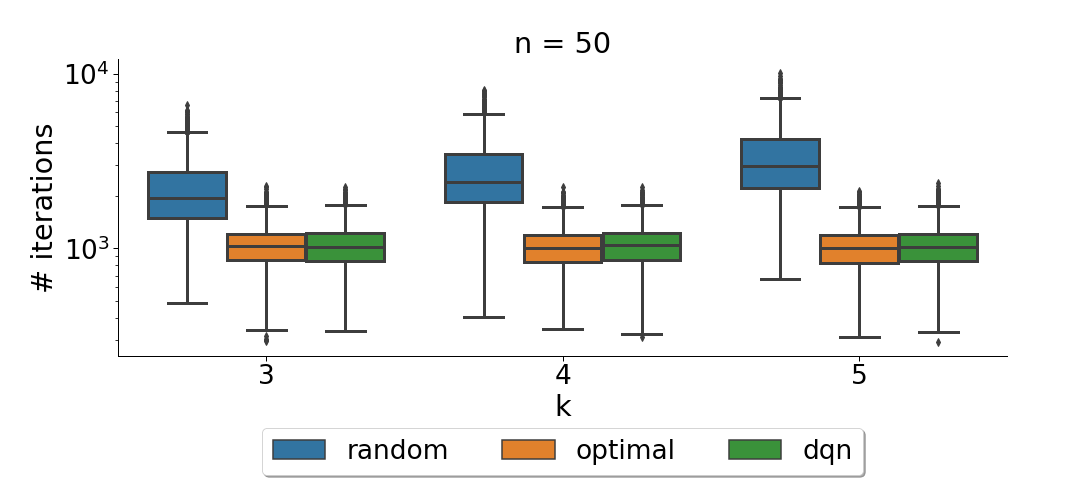}
    \caption{Performance of DDQN and optimal policies on three portfolio settings and three portfolio sizes ($n=50$). See also \Cref{sec:50}.}
    \label{fig:eval_n50_k456_three_settings}
    \Description[Short description.]{Long description.}
\end{figure}

\begin{figure}[t]
    \includegraphics[width=0.49\columnwidth]{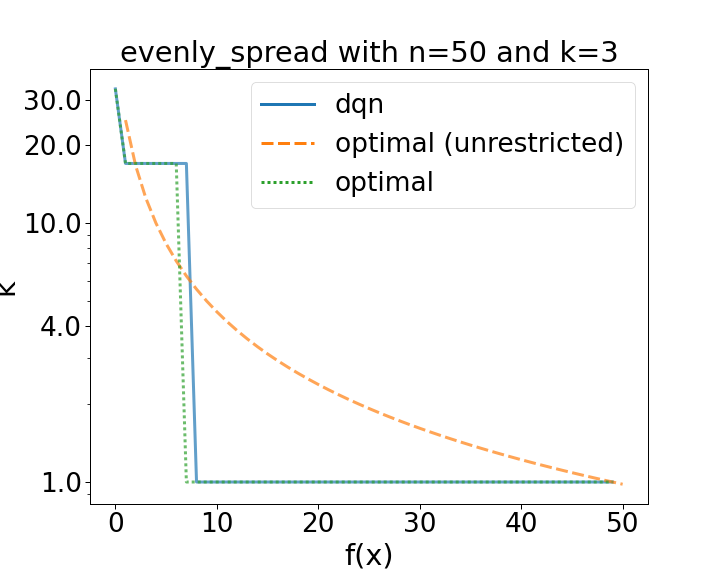}
    \includegraphics[width=0.49\columnwidth]{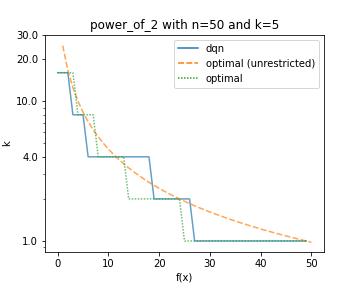}
    \caption{Two example DDQN best learned policies vs. the optimal policy for the same portfolio, and the optimal policy with unrestricted portfolio. See also \Cref{sec:50}.}
    \label{fig:example_policies}
    \Description[Short description.]{Long description.}
\end{figure}

As shown in \Cref{fig:eval_n50_k456_three_settings}, the performance of the DDQN policies is highly comparable to the optimal ones. DDQN is able to reach the performance of the optimal policy within $100\,000$ time steps in all cases. The learned policies are also quite similar to the optimal ones, with some slight discrepancy, as illustrated in \Cref{fig:example_policies}, where DDQN learned policies for two example settings (\evenlySpread with $k=3$, and \powersOfTwo with $k=5$).

\begin{figure}[t]
    \includegraphics[width=0.8\columnwidth]{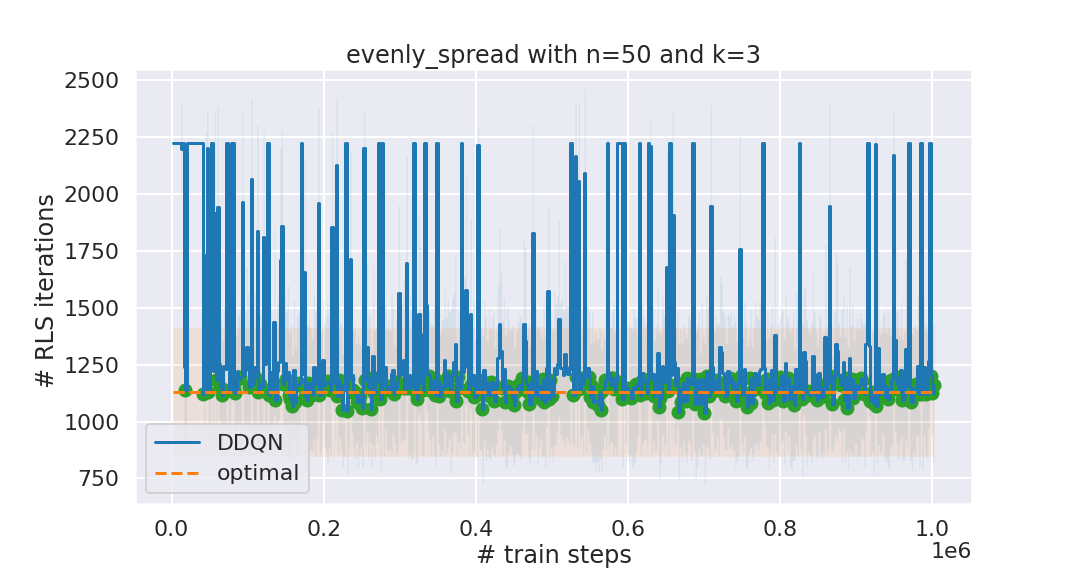}
    \includegraphics[width=0.8\columnwidth]{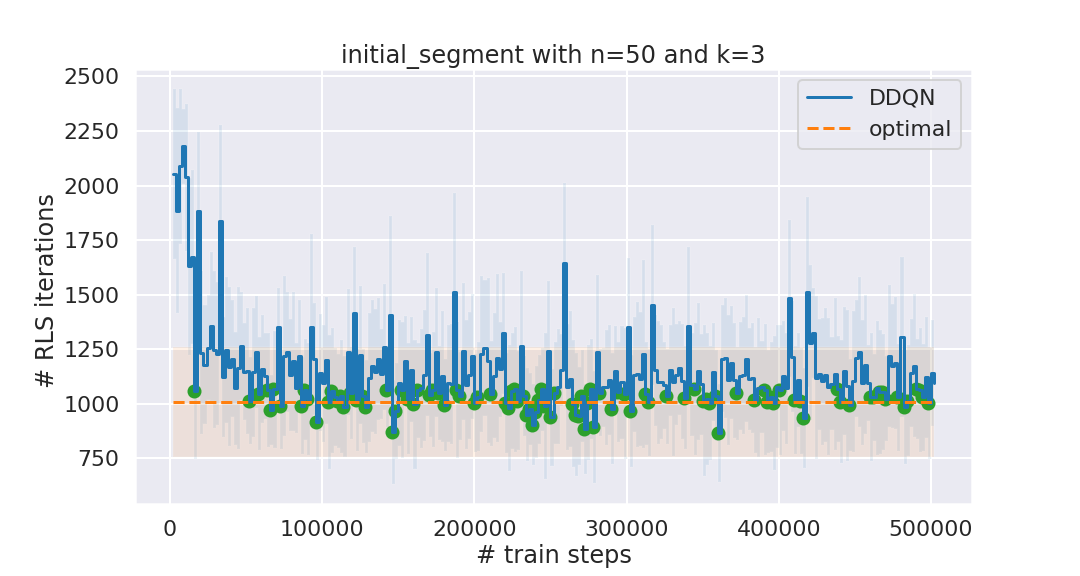}
    \caption{DDQN training progress with \evenlySpread and \initialSegment. The green dots are where the learned policies reach $0.25\,\%$ the standard deviation of the optimal policy's performance. See also \Cref{sec:50}.}
    \label{fig:ddqn_training_2_settings}
    \Description[Short description.]{Long description.}
\end{figure}

\begin{figure}[t]
    \centering
    \includegraphics[width=0.49\columnwidth]{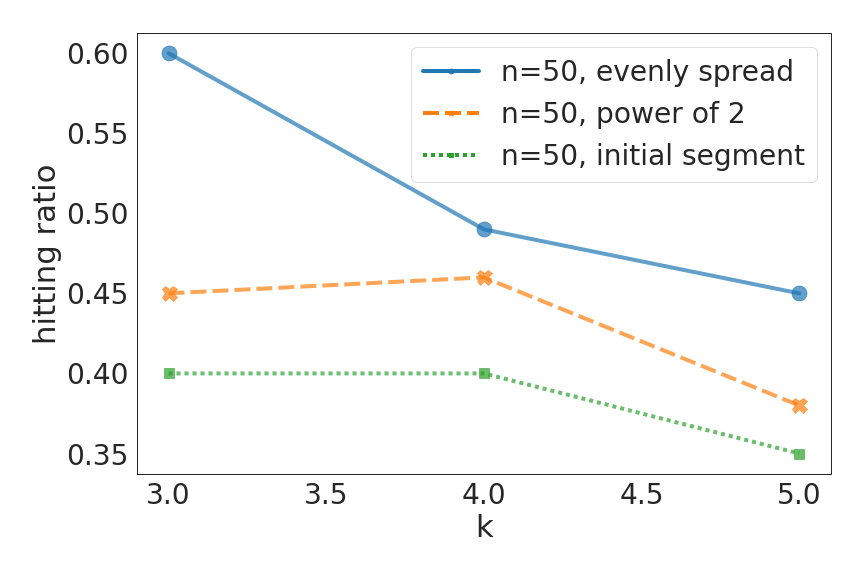}
    \includegraphics[width=0.49\columnwidth]{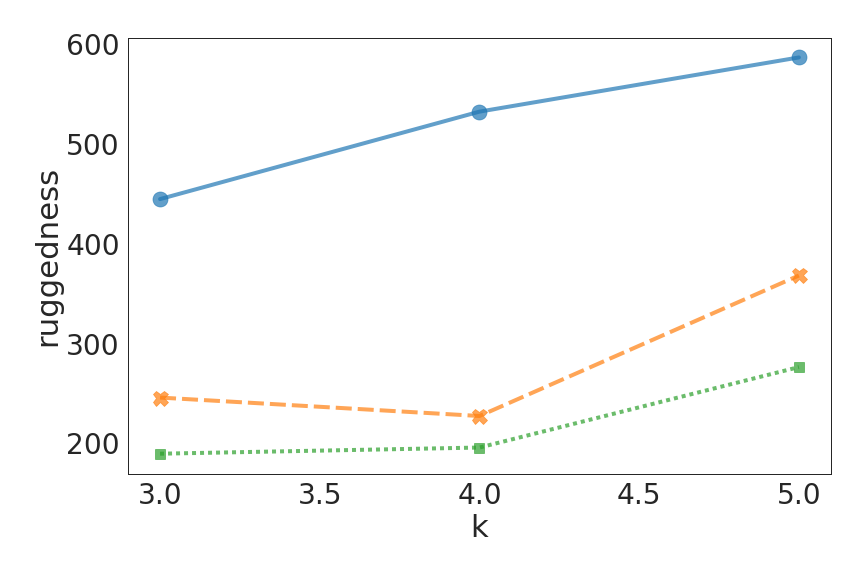}
    \caption{Hitting ratios and ruggedness of DDQN training progress for three portfolio settings ($n=50$). See also \Cref{sec:50}.}
    \label{fig:hitting_ratio_n50_k345_three_settings}
    \Description[Short description.]{Long description.}
\end{figure}

We now have a closer look at the training progress of each RL agent to see how different portfolio settings and portfolio sizes impact the learning behavior of DDQN.
To this end, we evaluate the learned policy during each DDQN training at every $2000$ time steps via $50$ runs and compare it with the optimal policy.
\Cref{fig:ddqn_training_2_settings} shows two example training progress plots of \evenlySpread and \initialSegment. Although DDQN frequently reaches the optimal area in both settings, there is a clear distinction between them: for \evenlySpread, DDQN's performance constantly jumps up and down with very high variance, while for \initialSegment, the performance progress is much smoother.
To quantify these properties of the training progress, we define two metrics for each DDQN training run: (i) \emph{hitting ratio} -- the frequency of evaluations in which the expected optimal performance is reached within $0.25\,\%$ of its standard deviation; and (ii) \emph{ruggedness} -- the standard deviation of performance difference between every pair of consecutively evaluated policies.
As shown in \Cref{fig:hitting_ratio_n50_k345_three_settings}, the RL agent gets the highest hitting ratios with \evenlySpread, followed by \powersOfTwo and \initialSegment. This can be explained due to the fact that the actions for \evenlySpread are very different from each other, some of which often perform very badly in general. Such differences can result in strong signals received by the agent during the training for distinguishing between good and bad policies, which can then help speed up the learning but also causes the landscapes to be less smooth (i.e., high  ruggedness) due to the large variance of performance between different policies. Similarly, the initial segment setting has the smallest difference between actions, and the RL agent has the lowest hitting ratios but smoother learning progress among the three settings.

\subsection{Analyzing the Impact of Portfolio Size and Problem Dimension}
\label{sec:dim}

\begin{figure}[t]
    \centering
    \includegraphics[width=0.8\columnwidth]{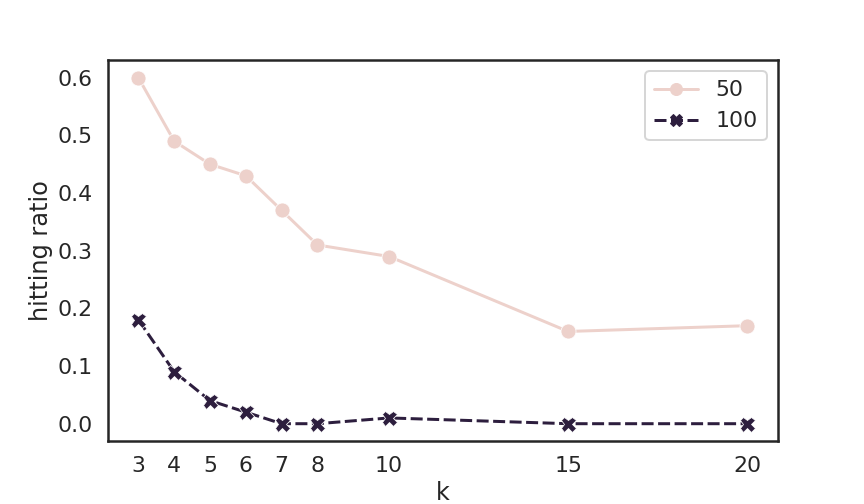}
    \caption{Hitting ratios of DDQN on \evenlySpread, with $n \in \{50, 100\}$ and $k \in \{3, 4, 5, 6, 7, 8, 10, 15, 20\}$. See also \Cref{sec:dim}.}
    \label{fig:hitting_ratio_evenly_spread_n50_n100}
    \Description[Short description.]{Long description.}
\end{figure}

\begin{figure}[t]
    \raggedright
    \includegraphics[width=\columnwidth]{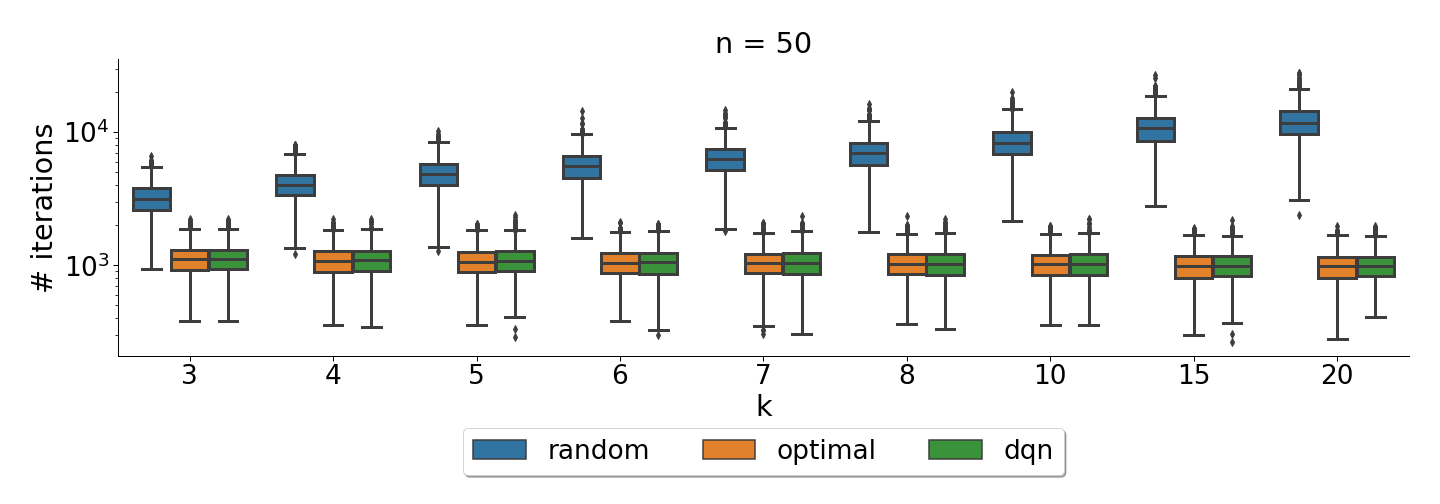}
    \includegraphics[width=1.1\columnwidth,]{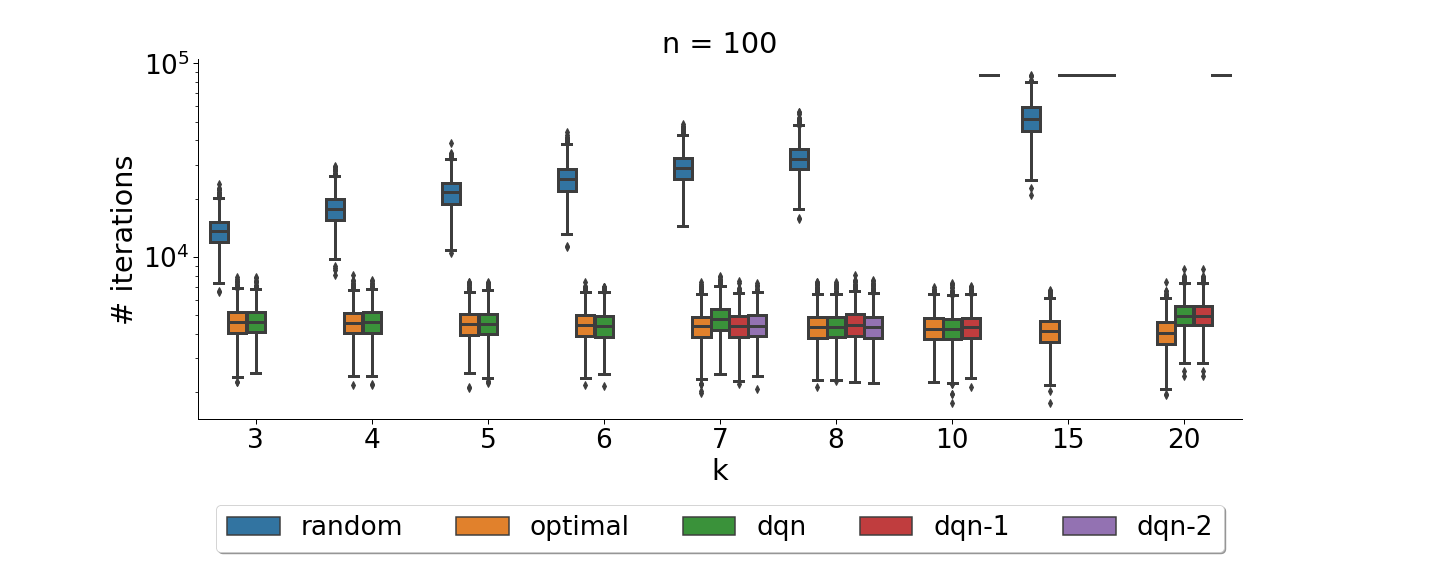}
    \caption{Performance of DDQN on \evenlySpread setting, with $k \in \{3, 4, 5, 6, 7, 8, 10, 15, 20\}$ and $n \in \{50, 100\}$. DDQN runs failing to learn are marked with a straight line. See also \Cref{sec:dim}.}
    \label{fig:performance_evenly_spread}
    \Description[Short description.]{Long description.}
\end{figure}

\Cref{fig:hitting_ratio_n50_k345_three_settings} indicates a strong relation between portfolio sizes and the learning ability of DDQN agents: the larger $k$ is, the smaller the hitting ratios. In the second set of experiments, we investigate further the impact of portfolio sizes and problem sizes on DDQN's learning behaviors.
We train DDQN agents on the evenly spread setting with a wider set of portfolio sizes $k \in \{3, 4, 5, 6, 7, 8, 10, 15, 20\}$ and with two problem sizes $n \in \{50, 100\}$. For $n=100$, we expect it to be more difficult for the RL agent to learn due to the larger episode lengths, therefore, the training budget is increased to $1.4$ million time steps.
As shown in \Cref{fig:hitting_ratio_evenly_spread_n50_n100}, DDQN hitting ratios decrease drastically as $k$ increases. For $n=100$ and $k \geq 7$, the hitting ratios are very close to zero. In fact, the performance of the learned policies by DDQN for $n=100$ and $k \in \{15, 20\}$ is no longer competitive to the optimal ones, as shown in \Cref{fig:performance_evenly_spread}. Looking into the detailed progress of each RL run, we find that for $k=7$, the agent barely hits the optimal policies (only $2$ times over $750$ evaluations), and for $k=15$, it has zero hitting rate.

The results so far indicate that we reach the learning limit of DDQN with the given setting. To confirm this hypothesis, we repeat the DDQN training two more times for each $k \geq 7$ and $n=100$. As shown in \Cref{fig:performance_evenly_spread}, for $n=100$ and all $k \geq 10$, there is at least one of three DDQN training runs where the agent does not learn anything, i.e., there is no progress in the entire training process.

\begin{figure}[t]
    \centering
    \includegraphics[width=0.8\columnwidth]{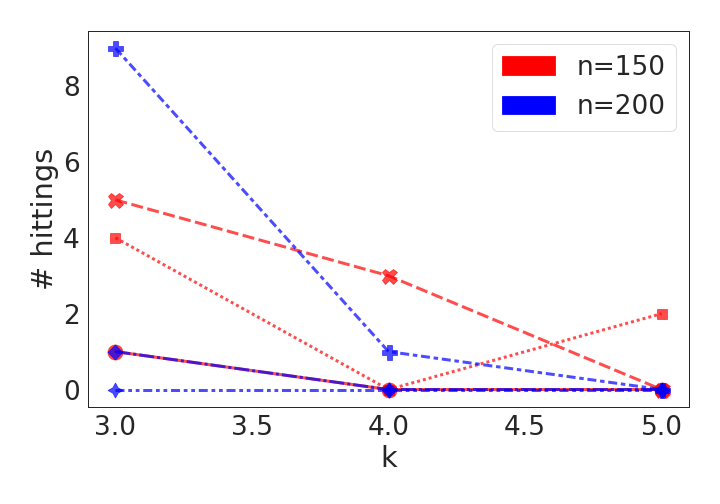}
    \caption{\#times DDQN reaches performance of the optimal policy on \evenlySpread, with $n \in \{150, 200\}$ and $k \in \{3, 4, 5\}$.
    Linestyles indicate individual runs with different seeds.
    See also \Cref{sec:dim}.
    }
    \label{fig:n_hits_n150_n200_k345}
    \Description[Short description.]{Long description.}
\end{figure}

In the last set of experiments, we investigate further the impact of problem dimension on the learning limit of DDQN. To this end, we train $3$ DDQN agents for each pair of $n \in \{150, 200\}$ and $k \in \{3,4,5\}$, with a budget of $1.4$ million steps and a walltime limit of $48$ hours. Within the given time limit, each DDQN agent can only reach $400\,000$ and $250\,000$ time steps for $n=150$ and $n=200$, respectively. This is due to the fact that the length of each evaluation episode increases quadratically with~$n$.
Figure~\ref{fig:n_hits_n150_n200_k345} shows the number of times each agent reaches the performance of the optimal policies during the entire training process. These results indicate that $n=200$ and $k=5$ is the final limit of our DDQN agent with the chosen hyperparameters, as neither of the three runs can get close to the optimal policy.

\balance
\section{Conclusion and Outlook}
\label{sec:conclusion}

We suggested the optimization of the \leadingones problem via the \oporls with fitness-dependent control policies as a benchmark problem in the context of dynamic algorithm configuration (DAC). This problem setting is theoretically very well understood, to the point that we could easily extend in this work the base case with full parameter portfolio $[1..n]$ to settings in which the search radii have to be chosen from a restricted portfolio $\K \subsetneq [1..n]$. That is, we can compute optimal control policies for any given combination of problem dimension $n$ and parameter portfolio $\K$. This allows us to create numerous problem instances of different size, which can be leveraged to gain structural insight into the behavior of DAC techniques. Empirically, we showed that DDQN efficiently learns optimal policies for the smaller \leadingones instances.
We also explored the settings at which DDQN with the chosen parameters and budget reaches its limits, in the sense that the learned policy is not close to optimal or even fails to learn entirely.

One way to overcome the limits of DDQN for larger problem and portfolio sizes could use AutoRL~\cite{AutoRLsurvey} to optimize its hyperparameters, such as the batch size, discounting factor, exploration strategy, choice of algorithm or network architecture.
Although it is known that RL agents are very brittle with respect to their hyperparameters, their influence on the learning algorithm is not well understood.
Our benchmark enables studying the effect of hyperparameters in a principled manner, which will potentially allow us to make RL agents more robust and easier to use in the context of dynamic algorithm configuration.
A favorable aspect in this context is that the evaluation times of the \leadingones benchmarks are very small, making a systematic investigation on the learning ability of RL agents computationally affordable.
In fact, we can reduce the evaluation times further if we replace the actual training process by a simulation that draws the rewards from the well understood reward distribution.

Since we understand the distribution of the reward function perfectly well, no matter the problem dimension, the state, nor the action that is played (\cref{eq:flipping_probability} essentially captures this information), we believe that it is feasible to extend recent theoretical investigations of classic (i.e., static) algorithm configuration~\cite{HallOS22} to the more general DAC setting.

Regarding the DAC setting, we did not exploit the full power of DAC in this work, as we trained and tested on the same problem instances and did not aim to derive policies that can be transferred to instances that were not part of the training set, as is classically done in algorithm configuration. Given the promising results of the DDQN agents, a reasonable next step would be to investigate the generalization ability of this approach with respect to problem dimension~$n$ or with respect to the portfolio~$\K$. Once established, the next step would then be to aim for generalizability across different problems, e.g., via a configurable benchmark generator that provides a good fit between problem representation and characteristics. The W-model constructions~\cite{WmodelInstancesASoC} could be a reasonable playground for first steps in this direction.
We note that generalization is an understudied topic in deep RL~\cite{kirk-arxiv21}, where DAC and our proposed benchmark can help to advance the field.

Another idea we are keen on exploring is to incorporate other state information into the policy of the \oporls than just the fitness.
For example, for \leadingones, \citet{BuzdalovB15CEC} show that adding information about the number of correct bits in the tail allows more efficient control policies. When considering a good configuration of DDQN, this approach could also be applied in order to derive approximately optimal policies for scenarios of state information for which no theoretical guarantees are known.

We emphasize that we investigated the new benchmarks for DAC only, but they are of course equally interesting for the parameter control setting. Techniques that model parameter control as a multi-armed bandit problem \cite[e.g.][]{DaCostaGECCO08,FialhoCSS10,DoerrDY16PPSN} can be straightforwardly applied to our benchmarks, as they typically require finite parameter portfolios anyway.
We also do not see greater obstacles to adjust other strategies, such as self-adaptive or self-adjusting parameter control mechanisms~\cite{EibenHM99}, even though the parameter encoding and update strategies may need to be redesigned to account for the restricted portfolio $\K$.

We hope that our work initiates a fruitful exchange of benchmarks between parameter control and dynamic algorithm configuration. With the growing literature on parameter control~\cite{KarafotiasHE15} and its theoretical analysis~\cite{DoerrD18chapter} we wish to provide other use-cases with a known ground truth. However, settings for which we have such detailed knowledge as for \leadingones are very rare. Even for \onemax, the ``drosophila of evolutionary computation''~\cite{FialhoCSS08}, the optimal mutation rates of the \oporls and the $(1+1)$~evolutionary algorithm are known only in approximate terms~\cite{DoerrDY20} or for specific problem dimensions~\cite{BuskulicD21,BuzdalovD20,BuzdalovD21}.
We believe that an active exchange of theoretically and automatically found policies will benefit both sides: empirical results may provide guidance or inspiration for theoretical analyses, whereas theoretical results can be used as benchmarks with ground truth, as we have demonstrated in this work.

\begin{acks}
Nguyen Dang is a Leverhulme Early Career Fellow.
André Biedenkapp and Frank Hutter acknowledge funding by the Robert Bosch GmbH.
This project has received funding from the European Union's Horizon 2020 research and innovation program under the Marie Skłodowska-Curie grant agreement No. 945298-ParisRegion\-FP. It is also supported by the Paris Île-de-France region, via the DIM RFSI AlgoSelect project and is partially supported by TAILOR, a project funded by EU Horizon 2020 research and innovation programme under GA No 952215.
The authors acknowledge the HPCaVe computing platform of Sorbonne Universit\'e for providing computational resources to this research project. The collaboration leading to this work was initiated at the 2020 Lorentz Center workshop ``Benchmarked: Optimization Meets Machine Learning''.
\end{acks}

\bibliographystyle{ACM-Reference-Format}

\end{document}